\documentclass[letterpaper, 10 pt, conference]{ieeeconf}  %

\IEEEoverridecommandlockouts                              %

\usepackage{amsmath}
\usepackage{amssymb}
\usepackage{amsfonts}
\usepackage{soul}
\usepackage[bookmarks=true]{hyperref}
\usepackage[normalem]{ulem}
\usepackage{tabularx} %
\usepackage{caption}
\usepackage{booktabs}
\usepackage{graphicx}
\usepackage{caption}
\usepackage{xspace}

\usepackage{subcaption}
\usepackage{booktabs} %
\usepackage[table]{xcolor} %
\newcommand{\dataaug}{ViRe}

\title{\LARGE \bf Concatenated Masked Autoencoders as Spatial-Temporal Learner}

\author{
Zhouqiang Jiang$^{1}$, Bowen Wang$^{2}$, Tong Xiang$^{1}$, Zhaofeng Niu$^{3}$, Hong Tang$^{4}$, Guangshun Li$^{3}$, Liangzhi Li$^{1*}$%
\thanks{$^{1}$Meetyou AI Lab, China.}
\thanks{$^{2}$Institute for Datability Science, Osaka University, Japan.}
\thanks{$^{3}$Department of Computer Science, Qufu Normal University, China.}
\thanks{$^{4}$Department of Information Engineering, East China Jiaotong University, China.}
\thanks{$^{*}$Corresponding author: {\tt \small liliangzhi@xiaoyouzi.com}}
}
\begin{document}

\sethlcolor{yellow}

\maketitle

\begin{abstract}

Learning representations from videos requires understanding continuous motion and visual correspondence between frames. In this paper, we introduce the Concatenated Masked Autoencoders (CatMAE) as a spatial-temporal learner for self-supervised video representation learning. For the input sequence of video frames, CatMAE keeps the initial frame unchanged while applying substantial masking (95\%) to subsequent frames. The encoder in CatMAE is responsible for encoding visible patches for each frame individually; subsequently, for each masked frame, the decoder leverages visible patches from both previous and current frames to reconstruct the original image. Our proposed method enables the model to estimate the motion information between visible patches, match the correspondence between preceding and succeeding frames, and ultimately learn the evolution of scenes. Furthermore, we propose a new data augmentation strategy, Video-Reverse (\dataaug), which uses reversed video frames as the model's reconstruction targets. This further encourages the model to utilize continuous motion details and correspondence to complete the reconstruction, thereby enhancing the model's capabilities. Compared to the most advanced pre-training methods, CatMAE achieves a leading level in video segmentation and action recognition tasks. Code is available at~\url{https://github.com/minhoooo1/CatMAE}.

\end{abstract}

\section{Introduction}

In the realm of image pre-training, Masked Autoencoders (MAE)~\cite{he2022masked} have demonstrated their effectiveness in learning visual representations by reconstructing missing patches from randomly masked input images. Recent research extends this paradigm to the pre-training for videos \cite{tong2022videomae,feichtenhofer2022masked,gupta2023siamese,dosovitskiy2020image}. Existing studies simply consider the spatio-temporal specificity of masking strategies in order to reduce inductive bias. However, they either overlook differences in spatial and temporal dimensions \cite{adelson1985spatiotemporal} or neglect the ability to model continuous motion. This leads to the dilemma where learned representations struggle to simultaneously incorporate both continuous motion information and the correspondence between video frames.

\begin{figure}[t]
    \centering
    \begin{subfigure}[b]{0.47\textwidth}
        \includegraphics[width=\textwidth]{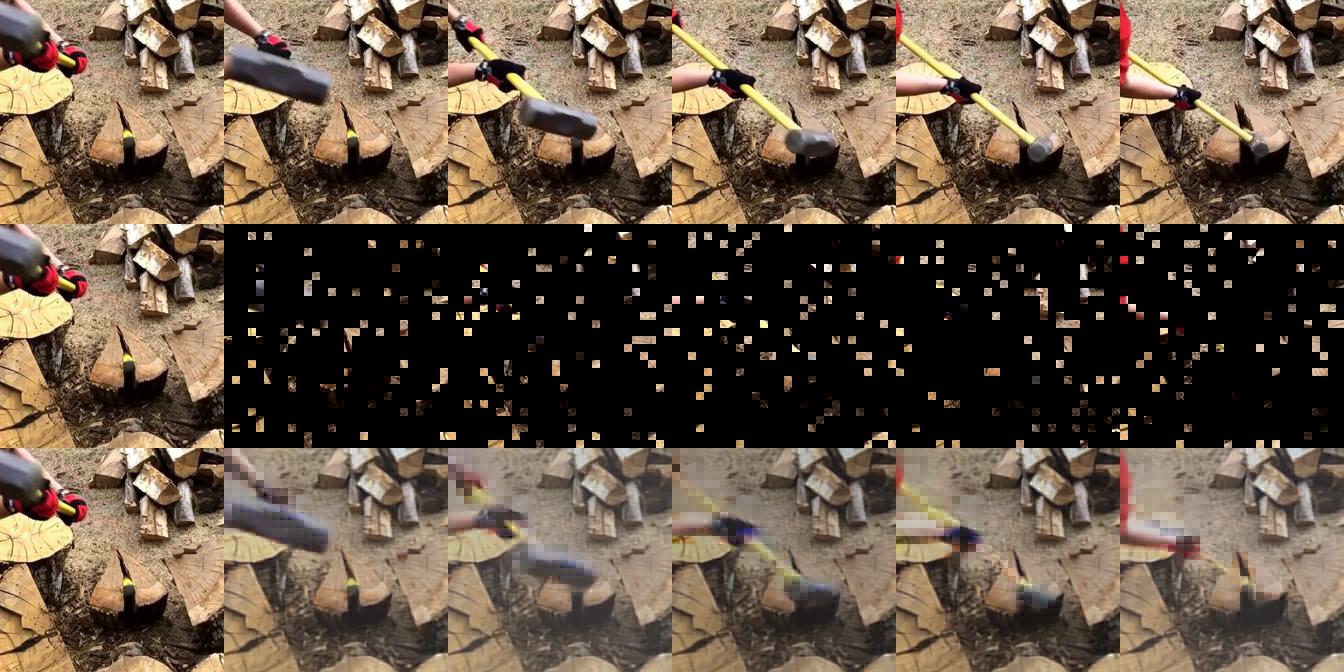}
        \caption{Chopping wood.}
        \label{fig:chopping-wood}
    \end{subfigure}
    \hfill
    \begin{subfigure}[b]{0.47\textwidth}
        \includegraphics[width=\textwidth]{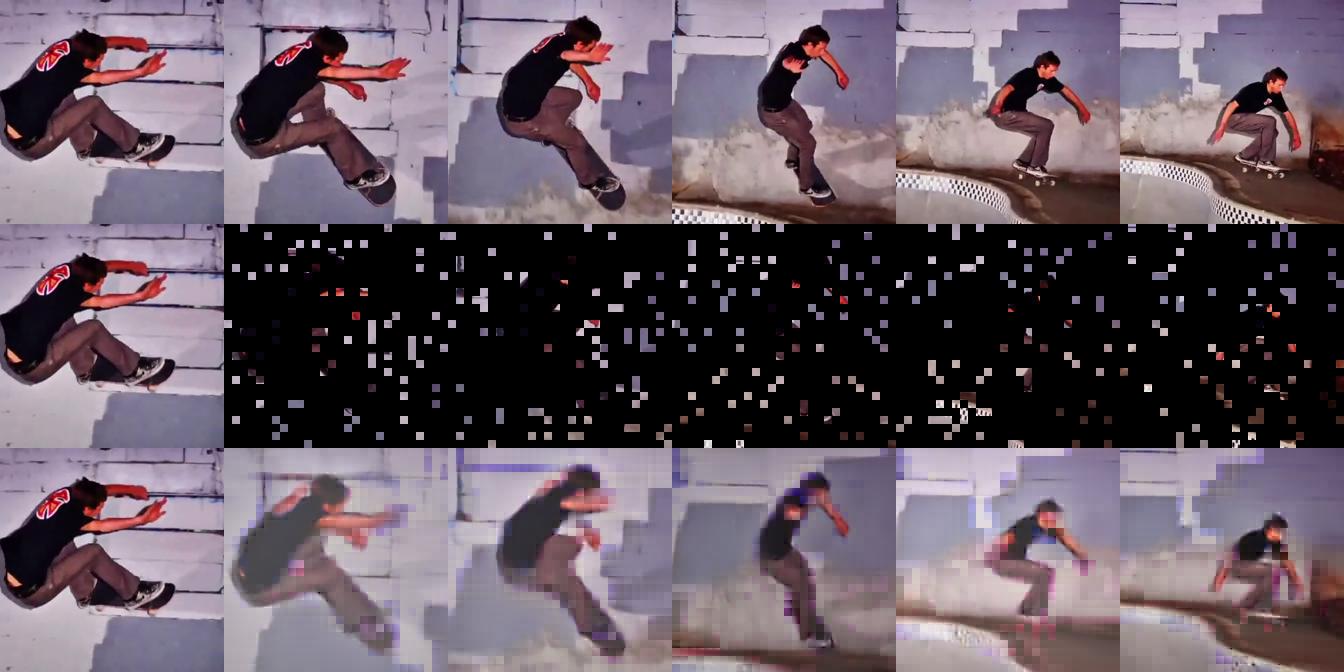}
        \caption{Skateboarding.}
        \label{fig:skateboarding}
    \end{subfigure}
    \caption{Visualizations on the Kinetics-400 \cite{carreira2017quo} validation set (masking rate 90\%). For each video sequence, we sample 6 frames with a frame gap of 4. Each subfigure displays the original frames (top), masked future frames (middle), and CatMAE reconstruction results (bottom). 
    }
    \label{fig:kinetics-visualizations}
    \vspace{-0.1in}
\end{figure}

Properly handling the temporal dimension in videos is essential to effectively utilize the masking and reconstruction paradigm. VideoMAE~\cite{tong2022videomae} and MAE-ST~\cite{feichtenhofer2022masked} expand 2D image patches into 3D cubes in videos, applying a self-supervised mask reconstruction pipeline on these cubes. However, the semantics of video frames vary slowly over time~\cite{zhang2012slow}, creating temporal redundancy which increases the risk of reconstructing missing cubes solely from the spatial-temporal neighborhood. Therefore, to prevent the model from learning shortcuts by exploiting this \textit{leaked information} during reconstruction, a very high mask rate is considered a straightforward universal solution, which also significantly reduces computational cost~\cite{feichtenhofer2022masked}. The pre-trained models from these methods demonstrate outstanding transferability on action recognition tasks.

Yet, utilizing 3D cube masking in the spatial-temporal dimension is sub-optimal for learning inter-frame correspondence; for instance, MAE even outperform VideoMAE in video segmentation tasks~\cite{pont20172017}. Therefore, SiamMAE~\cite{gupta2023siamese} proposes to use an asymmetric mask strategy for reconstructing future frames~\cite{lotter2016deep,mathieu2015deep} to learn inter-frame correspondence, and shows superiority in fine-grained correspondence tasks such as video object segmentation~\cite{pont20172017}, video part segmentation~\cite{zhou2018adaptive}, and pose tracking~\cite{jhuang2013towards}.

Continuous motion information and long-term correspondence usually span an extended period within a video. As such, predicting and reconstructing the future frames over a longer interval using the asymmetric masking strategy in SiamMAE is inherently hard due to their ambiguity. In extreme cases, the model may not be able to model any effective motion information, as the scenes in the two sampled frames could be entirely different.

To capture continuous motion information and long-term correspondence in videos, we propose \textbf{C}onc\textbf{at}enated \textbf{M}asked \textbf{A}uto\textbf{E}ncoders (CatMAE), which uses a concatenated information channel masking strategy to enhance the learning ability of the encoder. In our method, we first chronologically select $N$ frames from a video clip, keeping all patches of the first frame visible and then performing random masking on patches of subsequent frames with an extremely high mask ratio. The encoder encodes the visible patches from $N$ frames separately, and then the decoder reconstructs the masked patches for $N$-1 subsequent frames. Note that when reconstructing each frame's missing patches, the decoder utilizes cross-attention to receive information from the visible patches of previous and current frames. This concatenated information channel masking assists in modeling motion differences and correspondence, eventually estimating the long-term dynamic evolution of video frames. As shown in Fig~\ref{fig:kinetics-visualizations}, CatMAE is able to reconstruct a long video sequence. Furthermore, inspired by time's bidirectionality, we hypothesize that the reconstruction of reversed actions can reinforce the understanding of actions, and propose a video reverse data augmentation method, \textbf{Vi}deo-\textbf{Re}verse (\dataaug), to enhance the representation learning.

In conclusion, CatMAE enriches the exploration of the masking and reconstruction pipeline~\cite{he2022masked} in the field of self-supervised video representation learning. In video segmentation and action recognition tasks, the ViT \cite{dosovitskiy2020image} model pre-trained with CatMAE achieves leading results. Experiment results show that our proposed CatMAE and~\dataaug~arouse the full potential of paradigm of masking and reconstruction within the spatial-temporal domain.

\begin{figure*}[t] 
   \centering
   \includegraphics[width=\textwidth]{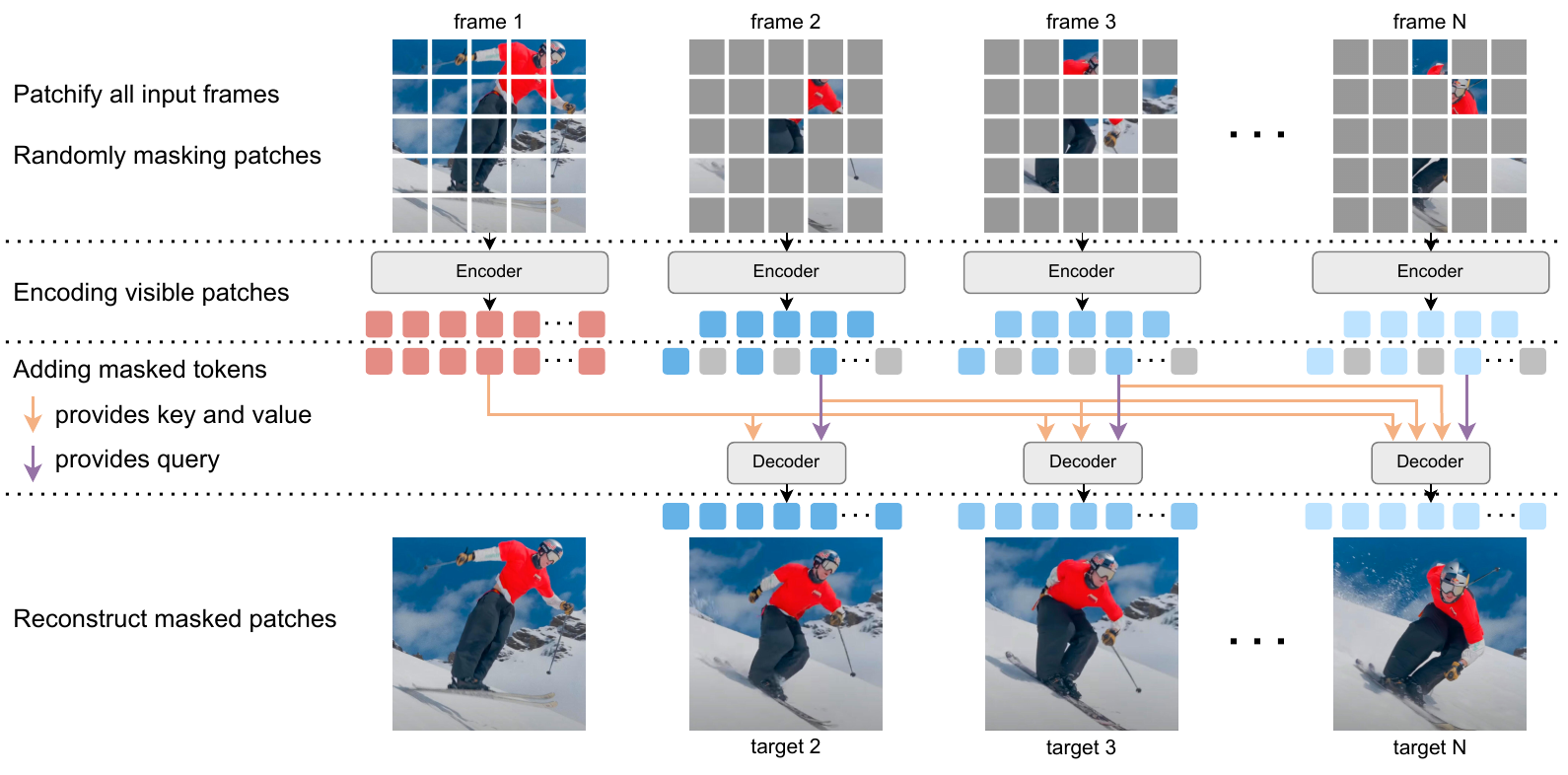} 
   \caption{Pipeline of our CatMAE. During pre-training, we chronologically extract $N$ frames from a video clip, keeping all patches of the first frame visible, and apply a very high masking ratio to mask the patches of the subsequent $N-$1 frames randomly. The visible patches of $N$ frames are independently processed by the ViT encoder. The decoder reconstructs the missing patches for the subsequent $N-$1 frames. Note that when reconstructing each frame's missing patches, the decoder receives information from all visible patches of previous frames through a sequence of cross-attention layers \cite{gupta2023siamese}.}
   \label{fig:arch}
   \vspace{-0.1in}
\end{figure*}

\section{Related Work}

\subsection{Masked Visual Modeling}

Masked visual modeling method \cite{vincent2008extracting} learn representations from images disrupted by masking. Essentially, they function as denoising autoencoders, disrupting input signals and reconstructing the original undamaged signals to learn effective representations. This paradigm has spawned a range of derivatives, such as reconstructing masking pixels \cite{vincent2010stacked, pathak2016context} or restoring lost color channels \cite{zhang2016colorful}. 

The success of masked language modeling \cite{devlin2018bert} in NLP's self-supervised pre-training, along with the popularity of ViT, has sparked extensive research into the use of transformer-based architectures for masked visual modeling within the field of computer vision \cite{bao2021beit, he2022masked, dong2023peco, wei2022masked, xie2022simmim, zhou2021ibot}. BEiT \cite{bao2021beit}, PeCo \cite{dong2023peco}, and iBOT \cite{zhou2021ibot} naturally inherit the idea of BERT \cite{devlin2018bert} and propose learning representations from images by predicting discrete tokens. 

Some other research \cite{chen2020generative, he2022masked, xie2022simmim} focuses on using pixels as prediction targets, which is simpler. MAE \cite{he2022masked} and SimMIM \cite{xie2022simmim}, by learning to reconstruct missing patches from randomly masked input image patches, achieve good visual representation. Moreover, MAE's encoder only handles visible patches under a high mask ratio, significantly speeding up training and achieving better transfer performance. MAE has also been straightforwardly extended to the video domain \cite{tong2022videomae, feichtenhofer2022masked}, reconstructing masked cubes. However, cube masking is sub-optimal for learning correspondence. SiamMAE \cite{gupta2023siamese} proposed an asymmetric masking strategy that retains past frames while reconstructing future frames containing a high proportion of masks, encouraging the model to model motion and match inter-frame correspondence. However, this strategy struggle to model long-term continuous motion information. Therefore, we propose a concatenated information channel masking reconstruction strategy that extends motion modeling to a theoretically infinite frame interval.

\subsection{Contrastive Based Self-supervision}

Effectively utilizing the temporal dimension is crucial in self-supervised video representation learning. Currently, there are various pretext tasks for pre-training, including predicting the future \cite{walker2016uncertain, gupta2022maskvit, vondrick2016anticipating}, segmenting pseudo ground truth \cite{pathak2017learning}, reconstructing future frames \cite{lotter2016deep, mathieu2015deep, srivastava2015unsupervised}, tracking \cite{wang2019learning, wang2015unsupervised}, reference coloring \cite{vondrick2018tracking}, and temporal ordering \cite{fernando2017self, lee2017unsupervised, misra2016shuffle, wei2018learning}. More advanced contrastive learning methods \cite{hadsell2006dimensionality, becker1992self} have been developed, which learn representations by modeling image similarities and dissimilarities \cite{he2020momentum, chen2020simple} or solely similarities \cite{chen2021exploring, grill2020bootstrap, caron2021emerging}. However, these methods rely on large batches \cite{chen2020simple}, multi-crops \cite{caron2020unsupervised}, negative key queues \cite{he2020momentum}, or custom strategies to prevent representation collapse \cite{grill2020bootstrap}. Their performance greatly depends on the choice of image augmentation \cite{chen2020simple}. In contrast, our method is based on a simple masking and reconstruction pipeline \cite{he2022masked}.

\section{Method}

In this section, We introduce each key component of CatMAE. The overall model architecture is shown in Fig~\ref{fig:arch}.

\noindent
\textbf{Patch Embedding.}\hspace{0.5cm}First, we chronologically select $N$ frames from a video clip as input sequence. The interval between frames is randomly chosen from a pre-determined frame gap range. Then, following the process of ViT~\cite{dosovitskiy2020image}, we divide each frame into non-overlapping patches. We apply linear projection and flatten the patches, concatenate a~\texttt{[CLS]} token to patches and eventually add position embedding to them.

\noindent
\textbf{Concatenated Information Channel Masking.}\hspace{0.5cm}We keep the first frame completely visible and apply masking with a high masking rate to $N-$1 subsequent frames, preserving only a minimal number of visible patches. Given that video signals are highly redundant \cite{zhang2012slow}, high masking rate prevents the model from utilizing information from adjacent frames during the reconstruction process, thus encouraging the encoder to capture motion information and correspondence. 

Visible patches in subsequent frames provide information channels, theoretically allowing the frames' sequence to be reconstructed indefinitely. To reconstruct the final frame, the model needs to reconstruct the continuous motion and correspondence of the intermediate frames, ultimately realizing the evolution between frames. (as shown in Fig.~\ref{fig:kinetics-visualizations}).

\noindent
\textbf{Encoder.}\hspace{0.5cm}We employ weight-shared vanilla ViT \cite{dosovitskiy2020image} to independently process all frames, with each encoder only encoding visible patches. This design significantly reduces temporal and memory complexity while achieving better performance \cite{he2022masked}. Furthermore, the patch embedding only adds a position embedding in the spatial structure, hence, the encoder is unaware of the temporal structure of the patches. However, the decoder's cross-attention layer needs to use the encoder's output to propagate the visible content. Even when the temporal structure is unknown, the encoder is forced to match the correspondence between spatial-temporal patches before and after motion. This correspondence assists the decoder in utilizing the information channels to propagate visible content, ultimately achieving the reconstruction of subsequent frames.

\noindent
\textbf{Decoder.}\hspace{0.5cm}We also employ a weight-shared decoder to separately predict the reconstructed frames' masked patches. Each decoder block consists of a cross-attention layer and a self-attention layer \cite{vaswani2017attention}. Specifically, the visible tokens of the frame to be reconstructed are projected via a linear layer and combined with mask tokens to form a set of full tokens, to which spatial position embedding is then added. Subsequently, the full tokens attend to all previously visible tokens (tokens of the first frame and other visible tokens) through a cross-attention layer, followed by mutual attention through a self-attention layer. Our concatenated information channel masking enables the decoder to enhance the encoder's ability to estimate motion offsets and learn correspondence. Finally, The output sequence of the decoder is used to predict the normalized pixel values in the masked patches \cite{he2022masked}, with an L2 loss applied between the decoder's prediction and ground truth.

\section{Experiments}

\subsection{Pre-training}
\noindent
\textbf{Architecture.}\hspace{0.5cm}
We use ViT-S \cite{touvron2021training} as the encoder, with a depth of 12 and a width of 384, which is similar to ResNet-50 \cite{he2016deep} in terms of the number of parameters (21M vs. 23M). The effectiveness of this setup is demonstrated in Table~\ref{tab:ablation-b} and~\ref{tab:ablation-c}.

\noindent
\textbf{Settings.}\hspace{0.5cm}
We conduct self-supervised pre-training on the Kinetics-400~\cite{carreira2017quo} action recognition dataset, which encompasses a total of 400 classes, comprising 240k training videos and 20k validation videos, and we only conduct pre-training on the training set. Considering the cost of training, we default to sampling three frames (224x224) from each video at certain frame intervals, and employ two image augmentation methods: RandomResizeCrop and horizontal flipping. A large portion (95\%) of the patches in the last two frames are randomly masked, and the batch size is 2048. The base learning rate is initialized to 1e-4 and follows a cosine schedule decay. We use the AdamW~\cite{loshchilov2017decoupled} optimizer, and the reconstruction losses for the second and third frames are scaled by 0.8 and 1.0, respectively. We adopt a repetition sampling \cite{gupta2023siamese, hoffer2020augment} factor of 2 and finally conduct pre-training for 300 effective epochs.

\subsection{Evaluation Method}
We conduct two downstream experiments, video segmentation and action recognition, to evaluate the correspondence and motion information of the obtained representations.

\noindent
\textbf{Fine-grained Correspondence Task}.\hspace{0.5cm}We use the semi-supervised video segmentation dataset DAVIS-2017 \cite{pont20172017} to evaluate the quality of representations in fine-grained correspondence tasks. We adopt the label propagation method used in \cite{jabri2020space, gupta2023siamese} to implement semi-supervised video segmentation. 
Given the ground truth labels of the first frame, we obtain k-nearest neighbor results through pixel feature similarity between frames. Afterward, the ground truth labels of the first frame are propagated to subsequent frames as pixel labels, based on the k-nearest neighbors' results. 

\noindent
\textbf{Action Recognition Task}.\hspace{0.5cm}We followed the evaluation protocol of VideoMAE \cite{tong2022videomae} for the motion evaluation on action recognition task by Kinetics-400 \cite{carreira2017quo}. In order to avoid a variety of meticulously crafted fine-tuning settings obstructing fair comparisons between different pre-training methods, we used the same pipeline to uniformly fine-tune all pre-trained models on the training set for 150 epochs and report the results on the validation set. To fairly compare with video MAEs \cite{tong2022videomae, feichtenhofer2022masked}, we also will extend the 2D patch embedding layer to a 3D embedding layer following the approach proposed in \cite{carreira2017quo}. 

\subsection{Main Results and Analysis}

\noindent
\textbf{Fine-grained Correspondence Task}.\hspace{0.5cm}Table~\ref{tab:davis-result} demonstrates all the methods using 480x880 images. We find
that CatMAE-ViT-S/8 surpasses all other self-supervised or supervised methods, and its performance is further enhanced when~\dataaug~is applied. This demonstrates that using reversed videos as reconstruction targets can further enhance the model's ability to match correspondence. As analyzed before, the cube masking recovery strategy cannot learn the inter-frame correspondence, resulting in MAE-ST and VideoMAE being significantly lower than CatMAE. In addition, CatMAE also significantly outperforms SiamMAE, owning a better segmentation performance with larger frame intervals (Table~\ref{tab:ablation-a}), proving that using the concatenated information channel masking strategy can capture longer and more accurate correspondence in videos.

\begin{table}[t]
\begin{tabular*}{\columnwidth}{@{\extracolsep{\fill}}lll|lcc}
\toprule
\textbf{Method} & \textbf{Backbone} & \textbf{Dataset} & \textbf{epoch} & $\mathcal{J}\&\mathcal{F}_m$ \\ \midrule
Surpervised \cite{he2016deep}           & ResNet-18     & ImageNet  &  -   & 63.8     \\
Surpervised \cite{he2016deep}           & ResNet-50     & ImageNet  & -    & 67.9     \\ \midrule
SimSiam \cite{chen2021exploring}        & ResNet-50     & ImageNet  & 100  & 66.7     \\ 
MoCo-v2 \cite{chen2020improved}         & ResNet-50     & ImageNet  & 800  & 66.5     \\
DINO \cite{caron2021emerging}           & ResNet-50     & ImageNet  & 800  & 66.1     \\
VFS \cite{xu2021rethinking}             & ResNet-50     & Kinetics  & 500  & 65.0     \\ 
CRW \cite{jabri2020space}               & ResNet-18     & Kinetics  & -    & 67.4      \\  \midrule
MAE \cite{he2022masked}                 & ViT-B/16      & ImageNet  & 1600 & 52.0      \\
MAE-ST \cite{feichtenhofer2022masked}   & ViT-L/2x16x16 & Kinetics  & 100  & 54.6      \\
VideoMAE \cite{tong2022videomae}        & ViT-S/2x16x16 & Kinetics  & 1600 & 39.3      \\ \midrule
DINO \cite{caron2021emerging}           & ViT-S/16      & ImageNet  & 800  & 63.0   \\
SiamMAE \cite{gupta2023siamese}         & ViT-S/16      & Kinetics  & 400  & 61.3     \\ 
CatMAE                                  & ViT-S/16      & Kinetics  & 300  & 62.0     \\
CatMAE +~\dataaug                       & ViT-S/16      & Kinetics  & 300  & 62.5     \\ \midrule
DINO \cite{caron2021emerging}           & ViT-S/8       & ImageNet  & 800  & 70.0   \\
SiamMAE \cite{gupta2023siamese}         & ViT-S/8       & Kinetics  & 400  & 69.0    \\
CatMAE                                  & ViT-S/8       & Kinetics  & 300  & 69.5    \\
CatMAE +~\dataaug                       & ViT-S/8       & Kinetics  & 300  & \textbf{70.4}    \\
\bottomrule
\end{tabular*}
\caption{Video object segmentation results on DAVIS 2017 validation set. Comparison of our method (\dataaug~ indicates that~\dataaug~is utilized during the pre-training phase), with the state-of-the-art self-supervised methods and pre-trained feature baselines. $\mathcal{F}$ is a boundary alignment metric, while $\mathcal{J}$ measures region similarity as IOU between masks.}
\label{tab:davis-result}
\end{table}

\begin{table}[t]
\begin{tabular*}{\columnwidth}{@{\extracolsep{\fill}}lll|ll}
\toprule
\textbf{Method}                              & \textbf{Backbone}        & \textbf{Epoch}  & \textbf{Top-1} & \textbf{Top-5} \\
\midrule

VideoMAE~\cite{tong2022videomae}    & ViT-S/2x16x16      & 10   & 47.70            &  73.99         \\ 
SiamMAE~\cite{gupta2023siamese}     & ViT-S/16        & 10   & 40.68            &  67.39     \\
CatMAE                              & ViT-S/16        & 10   & 46.41            &  72.87      \\ \midrule

VideoMAE~\cite{tong2022videomae}    & ViT-S/2x16x16      & 150   & 72.54            &  89.60        \\ 
SiamMAE~\cite{gupta2023siamese}     & ViT-S/16        & 150   & 66.42            &  85.96     \\
CatMAE                              & ViT-S/16        & 150   & 68.50           &  87.63         \\
\bottomrule
\end{tabular*}
\caption{Action recognition results during the fine-tuning stage on the Kinetics-400 validation set. Comparison of our method, with MAE variants.}
\label{tab:kinetics-result}
\end{table}

\noindent
\textbf{Action Recognition Task}.\hspace{0.5cm}
We implement this experiment with two competitive video representation methods VideoMAE and SiamMAE. CatMAE, SiamMAE, and VideoMAE respectively undergo pretraining for 300, 400, and 1600 epochs. In Table~\ref{tab:kinetics-result}, it is observed that during the early stage of fine-tuning, CatMAE exhibits a significantly stronger transfer capability than SiamMAE, with an enhancement in top-1 accuracy by 5.73\%. This indicates that our proposed concatenated information channel masking reconstruction pre-training strategy can capture more motion information than the asymmetric mask reconstruction strategy of SiamMAE. Even when the fine-tuning saturates, CatMAE still leads SiamMAE by 2.08\% in top-1 accuracy. However, the gap between CatMAE and VideoMAE widen, which we believe mainly stems from the method of modeling tokens. VideoMAE directly models token embedding using a 3D cube, which essentially makes VideoMAE a type of 3D network. VideoMAE naturally contains more motion information than CatMAE which encodes 2D patch as tokens. Yet, in the initial stage of fine-tuning, CatMAE manages to maintain comparable transfer ability to VideoMAE (only 1.29\% lower in top-1 accuracy), performing far better than the SiamMAE (7.02\% higher). Our CatMAE maximally exploits the motion information on the 2D patch. %

\begin{table*}[t]
    \centering
    \begin{subtable}[t]{0.32\textwidth}
        \centering
        \begin{tabular}{ccll}
        rec. num    & $\mathcal{J}\&\mathcal{F}_m$ & $\mathcal{J}_m$ & $\mathcal{F}_m$ \\ \toprule
             1      & 54.3    & 52.7   & 55.9   \\
             2      & \cellcolor{gray!25}56.0    & \cellcolor{gray!25}54.2   & \cellcolor{gray!25}57.8   \\
             3      & 57.5    & 55.3   & 59.7  \\
             4      & \textbf{58.2}    & \textbf{56.2}   & \textbf{60.2}  
        \end{tabular}
        \caption{\textbf{Reconstruction number}. More reconstructions can improve performance.}
        \label{tab:ablation-a}
    \end{subtable}\hfill
    \begin{subtable}[t]{0.32\textwidth}
        \centering
        \begin{tabular}{ccll}
        dim & $\mathcal{J}\&\mathcal{F}_m$ & $\mathcal{J}_m$ & $\mathcal{F}_m$ \\ \toprule
        192 & \cellcolor{gray!25}56.0   & \cellcolor{gray!25}54.2   & \cellcolor{gray!25}57.8   \\
        256 & \textbf{59.9}   & \textbf{57.4}   & \textbf{62.5}   \\
        512 & 57.3   & 55.0   & 59.6   \\
        1024& 55.6   & 53.2   & 58.0
        \end{tabular}
        \caption{\textbf{Decoder width.} A narrower decoder can achieve the best performance.}
        \label{tab:ablation-b}
    \end{subtable}\hfill
    \begin{subtable}[t]{0.32\textwidth}
        \centering
        \begin{tabular}{ccll}
        blocks & $\mathcal{J}\&\mathcal{F}_m$ & $\mathcal{J}_m$ & $\mathcal{F}_m$ \\ \toprule
        1   & 51.7   & 49.7   & 53.7   \\
        2   & \cellcolor{gray!25}56.0   & \cellcolor{gray!25}54.2   & \cellcolor{gray!25}57.8   \\
        4   & \textbf{60.0}   & \textbf{57.3}   & \textbf{62.6} \\
        8   & 57.4   & 54.0   & 60.8
        \end{tabular}
        \caption{\textbf{Decoder depth.} 4 blocks of
decoder achieve the best performance.}
        \label{tab:ablation-c}
    \end{subtable}
    
    \vspace{0.5cm}

    \begin{subtable}[t]{0.32\textwidth}
        \centering
        \begin{tabular}{ccll}
        mask ratio & $\mathcal{J}\&\mathcal{F}_m$ & $\mathcal{J}_m$ & $\mathcal{F}_m$ \\ \toprule
        0.90, 0.90     & 54.5     & 52.8      & 56.2  \\
        0.95, 0.90     & 54.6     & 53.0      & 56.2  \\
        0.90, 0.95     & 55.6     & 53.7      & 57.5 \\
        0.95, 0.95     & \cellcolor{gray!25}\textbf{56.0}    & \cellcolor{gray!25}\textbf{54.2}     & \cellcolor{gray!25}\textbf{57.8}   \\
        0.99, 0.99     &  48.3    &   46.0    &  50.6
        \end{tabular}
        \caption{\textbf{Mask ratio}. A high masking ratio (95\%) works well, but not to extremes.}
        \label{tab:ablation-d}
    \end{subtable}\hfill
    \begin{subtable}[t]{0.32\textwidth}
        \centering
        \begin{tabular}{ccll}
        scale & $\mathcal{J}\&\mathcal{F}_m$ & $\mathcal{J}_m$ & $\mathcal{F}_m$ \\ \toprule
        0.5 1.0     & 54.6    & 52.5   & 56.7  \\
        0.8 1.0     & \cellcolor{gray!25}\textbf{56.0}    & \cellcolor{gray!25}\textbf{54.2}   & \cellcolor{gray!25}\textbf{57.8}   \\
        1.0 0.8     & 55.0    & 53.4   & 56.7   \\
        1.0 1.0     & 55.8    & 54.0   & 57.6    \\
        &           &          &\\
        \end{tabular}
        \caption{\textbf{Loss scale.} slightly emphasizing the reconstruction of the third frame is optimal.}
        \label{tab:ablation-e}
    \end{subtable}\hfill
    \begin{subtable}[t]{0.32\textwidth}
        \centering
        \begin{tabular}{ccll}
        frame gap & $\mathcal{J}\&\mathcal{F}_m$ & $\mathcal{J}_m$ & $\mathcal{F}_m$ \\ \toprule
        {[4,16]-[4,16]}    & \cellcolor{gray!25}56.0 & \cellcolor{gray!25}54.2  & \cellcolor{gray!25}57.8   \\
        {[4,48]-[4,16]}    & 55.2 & 53.3  & 57.1    \\
        {[4,16]-[4,48]}    & \textbf{56.8} & \textbf{54.9}  & \textbf{58.6}     \\
        {[4,48]-[4,48]}    & 55.6 & 53.9  & 57.3 \\
        &           &          &\\
        \end{tabular}
        \caption{\textbf{Frame gap.} Using a smaller sample rate between the first and 2nd frames works better.}
        \label{tab:ablation-f}
    \end{subtable}\hfill
    \caption{CatMAE ablation experiments with ViT-S/16 on DAVIS 2017 validation set. If not specified, the default is: the number of reconstructions is 2, the decoder has a width of 192 and a depth of 2, the Mask ratio and Loss scale for the two reconstruction frames are [0.95, 0.95] and [0.8, 1.0], respectively, the frame gap between frames 1-2 and 2-3 are both [4,16]. Default settings are marked in \colorbox{gray!25}{grey}.}
    \label{tab:ablation}
\end{table*}    

\begin{figure*}[t]
    \centering
    \includegraphics[width=\textwidth]{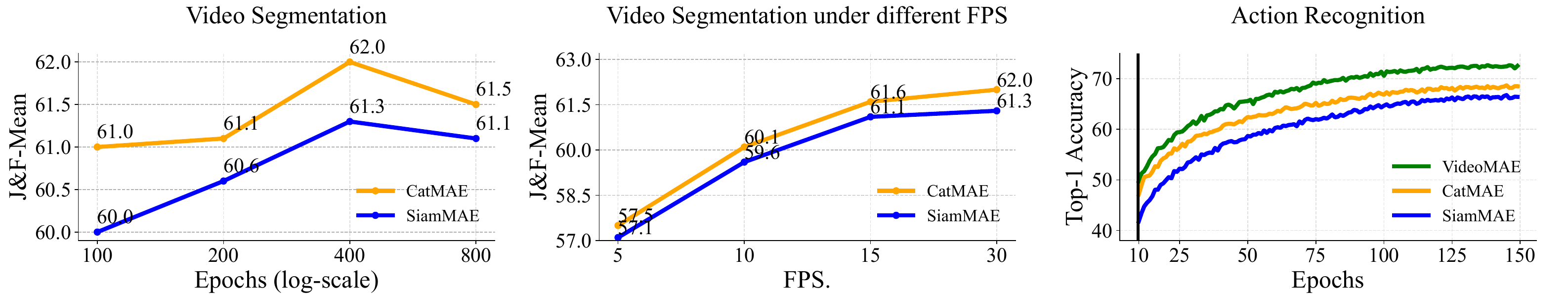}
    \caption{Impact of pretraining epochs and propagation FPS on video segmentation. Comparison of Top-1 accuracy for VideoMAE, CatMAE, and SiamMAE under different fine-tuning epochs.}
    \label{fig:seg-action-fps}
    
\end{figure*}

\subsection{Ablation Experiments}
We conduct an ablation experiment to understand each component's contribution. Details can be found in Table~\ref{tab:ablation}.

\noindent
\textbf{Number of Reconstructions.}\hspace{0.5cm}
Fig.~\ref{fig:kinetics-visualizations} shows the model's reconstruction results for future frames. As can be seen, the model is highly sensitive to motion elements in video frames. In Fig.~\ref{fig:chopping-wood}, the model can distinguish between static backgrounds and accurately reconstruct the motion of waving hammers. Beyond its motion modeling capability, the model can copy the background from the first frame and paste it into future frames when dealing with static backgrounds (such as the static woodpile). 
As shown in Table~\ref{tab:ablation-a}, the video segmentation performance steadily improves with the increase in the number of reconstructions. When the number of reconstructions is 4, the performance is higher in comparison with the number being 1 ($\uparrow$3.9\%). This validates the benefits of using a concatenated information channel masking strategy to extend multi-frame reconstruction.

\begin{figure*}
    \centering
    \begin{subfigure}[b]{0.24\textwidth}
        \includegraphics[width=\textwidth]{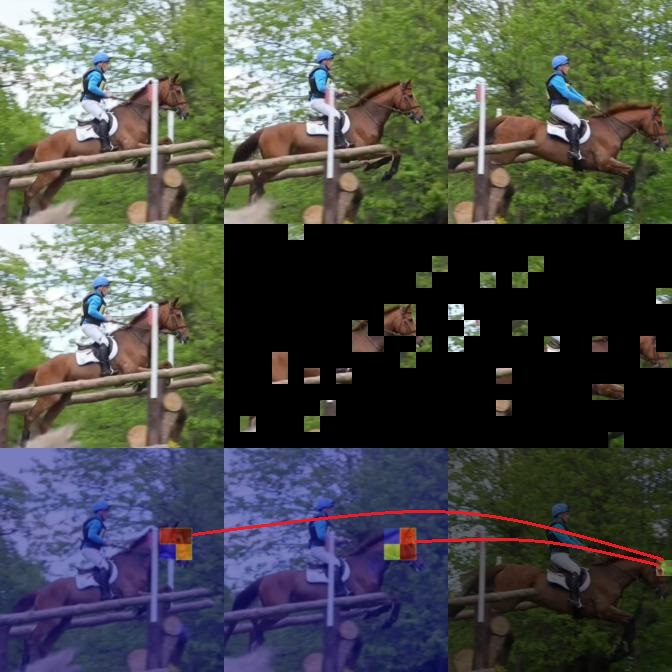}
        \caption{Horse.}
        \label{fig:cross-attention-horse}
    \end{subfigure}
    \hfill 
    \begin{subfigure}[b]{0.24\textwidth}
        \includegraphics[width=\textwidth]{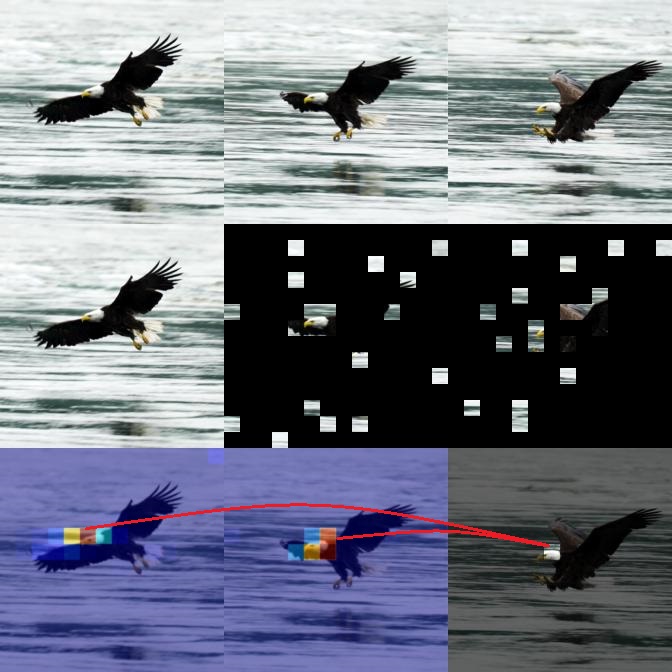}
        \caption{Seabird.}
        \label{fig:cross-attention-seabird}
    \end{subfigure}
    \hfill 
    \begin{subfigure}[b]{0.24\textwidth}
        \includegraphics[width=\textwidth]{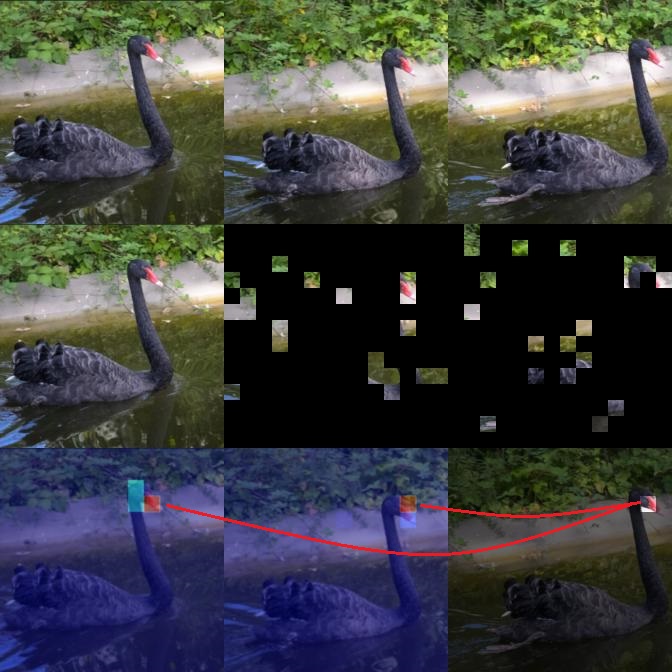}
        \caption{Black swan.}
        \label{fig:cross-attention-swan}
    \end{subfigure}
    \hfill
    \begin{subfigure}[b]{0.24\textwidth}
        \includegraphics[width=\textwidth]{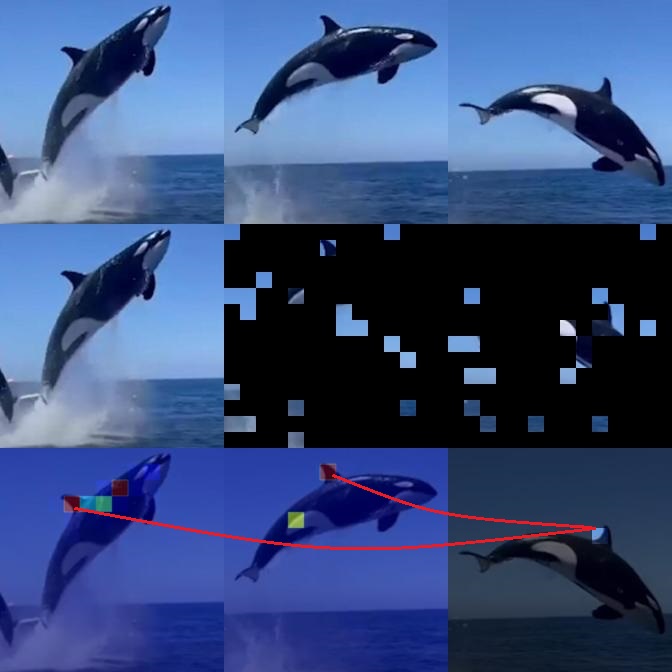}
        \caption{Orca.}
        \label{fig:cross-attention-orca}
    \end{subfigure}
    
    \caption{Cross-attention maps from a ViT-S/16 decoder (samples from DAVIS-2017 \cite{pont20172017}). We visualize the attention of the mask patch to be reconstructed in the third frame towards the visible patches in the first and second frames at the average head in the last layer of the decoder.}
    \label{fig:cross-attention}
    \vspace{-0.1in}
\end{figure*}

\noindent
\textbf{Decoder Capacity.}\hspace{0.5cm}
In Table~\ref{tab:ablation-b} and~\ref{tab:ablation-c}, we respectively examine the width and depth of the decoder. We found that relative to the encoder (384-d, 12-block), a smaller decoder (256-d, 4-block) is critically important for fine-grained correspondence tasks. This can be explained by the relationship between pixel reconstruction and fine-grained feature correspondence: during pixel reconstruction, the encoder is sensitive to motion offsets, thus obtaining a more reasonable correspondence representation. However, when the decoder capacity is too large, it weakens the encoder's sensitivity to motion. Therefore, in terms of the width and depth settings for the decoder, an optimal width of 256 is 4.3\% better than the maximum width of 1024 (59.9\% vs. 55.6\%). Additionally, an optimal depth of 4 is 2.6\% better than the depth of 8 (60.0\% vs. 57.4\%). These optimal decoder capacities are much smaller than the encoder, underscoring the importance of utilizing a smaller decoder.

\noindent
\textbf{Mask Ratio.}\hspace{0.5cm}
In Table~\ref{tab:ablation-d}, we compare different mask ratio. Based on previous experience with video MAEs, we default to ablating from a high mask ratio (90\%), but under such a high mask ratio, increasing the mask ratio of the last two frames to 95\% still shows performance improvement (54.5\% vs. 56.0\%). However, when using an extreme mask ratio of 99\% (Only one patch is visible), the performance drops sharply. An information channel with limited capacity is crucial for the model to learn correspondence. However, when the channel capacity is reduced to its extreme limit, it becomes difficult for the model to match correspondence. Hence, we default to using a mask ratio of 95\%.

\noindent
\textbf{Loss Scale.}\hspace{0.5cm}
In Table~\ref{tab:ablation-e}, we compared the reconstruction loss scale between the second and third frames. We found that slightly emphasizing the reconstruction of the third frame is better than that of the second (compare the second and third rows in Table~\ref{tab:ablation-e}). We believe that focusing more on the third frame's reconstruction facilitates the progressive transmission of continuous motion information to the final frame. However, neglecting the reconstruction of the second frame by reducing its loss scale to 0.5 leads to performance degradation. This might weaken the model's capability to capture the intermediate motion, thus affecting the propagation of continuous motion information.

\noindent
\textbf{Frame Gap.}\hspace{0.5cm}
In Table~\ref{tab:ablation-f}, we compare different frame gaps. While a larger frame gap sampling can encompass richer motion information, reconstructing two frames with large gaps does not bring performance improvement to the model (compare the first and fourth rows). This may be because when the frame gap becomes too large, the scene changes are too dramatic, which hinders the model's ability to model motion information. Additionally, we found that having a smaller gap between the first and second frames and a larger gap between the second and third frames (see the third row in Table) performs better than the reverse configuration (see the second row in Table) (56.8\% vs. 55.2\%). This observation can also be interpreted as the propagation of continuous motion being hindered: a large gap between the first and second frames makes capturing intermediate motion difficult, thereby blocking the transmission of continuous motion information within the information channels.

\noindent
\textbf{Training Epochs and Propagation FPS.}\hspace{0.5cm}
In Fig. \ref{fig:seg-action-fps}, we analyze the impact of training epochs and propagation FPS on video segmentation, discovering that optimal performance is achieved at pre-training 400 epochs, after which the performance gradually declines, possibly due to overfitting caused by small decoder.  We observe that under various low FPS conditions, CatMAE consistently outperforms SiamMAE, demonstrating that CatMAE possesses a stronger ability to match long-term correspondence. Additionally, we delve into the influence of fine-tuning epochs. CatMAE initially exhibits robust transferability but the gap widens with VideoMAE over time due to differences in token embedding. Simplistically speaking, CatMAE can be seen as a "additional frame reconstruction" version of SiamMAE. Therefore, according to Action Recognition figure, we estimate that using CatMAE to pre-train weights for reconstructing future four frames holds promise for reaching the fine-tuning performance of VideoMAE.

\subsection{Motion and Correspondence Analysis}
In Fig.~\ref{fig:cross-attention}, we visualize the cross-attention map averaged across different heads in the final layer of the ViT-S/16 decoder. Specifically, we input the first frame and the subsequent two frames with masks, followed by reconstructing the third frame. We collect the cross-attention from the mask patch in the third frame towards the visible patches in the first and second frames and then visualize this attention map. We find that the model is sensitive to motion, focusing on the correspondence between patches under continuous motion. For instance (Fig.~\ref{fig:cross-attention-horse}), when calculating the correspondence of the horse's head patch in the third frame, the model can clearly distinguish the semantics of the brown region and attention accurately focused on the head of the horse in previous frames.

Even when the background color is similar to the possible corresponding area (such as the white head of the water bird and the reflective water surface Fig.~\ref{fig:cross-attention-seabird}), the model's corresponding capability remains robust. Certainly, the model can also establish correspondence by leveraging unique color features (such as the red bill of a swan Fig.~\ref{fig:cross-attention-swan}) or specific semantic structures (such as the fin of an Orca Fig.~\ref{fig:cross-attention-orca}). These provide direct evidence of the model's ability to estimate motion and learn correspondence.

\section{Conclusion}
In this paper, we propose CatMAE for self-supervised video representation learning. It leverages a concatenated information channel masking strategy to address the limitations posed by cube masking and enhances the capability to capture long-term and corresponding motion compared to asymmetric masking. Our results show that it is competitive in comparison with state-of-the-art methods across both video segmentation and action recognition tasks. One distinctive feature of our training pipeline is the propagation of reconstruction information from the initial frame throughout the entire video sequence. This theoretically unlimited propagation showcases CatMAE's potential to learn long-term video representations. Our future work focuses on extending the application of CatMAE to real-world scenarios involving embodied agents, such as robots.

\newpage
\bibliography{refs} %
\end{document}